\documentclass{article}
\usepackage{ijcai23}

\usepackage{times}
\usepackage{soul}
\usepackage{url}
\usepackage[hidelinks]{hyperref}
\usepackage[utf8]{inputenc}
\usepackage[small]{caption}
\usepackage{graphicx}
\usepackage{amsmath}
\usepackage{amsthm}
\usepackage{booktabs}
\usepackage{algorithm}
\usepackage{algorithmic}
\usepackage[switch]{lineno}

\usepackage{enumitem}
\usepackage{adjustbox}

\newcommand{\citeal}[1]{\citeauthor{#1}~[\citeyear{#1}]}

\newcommand\blfootnote[1]{%
\begingroup 
\renewcommand\thefootnote{}\footnote{#1}%
\addtocounter{footnote}{-1}%
\endgroup 
}

\usepackage{tikz}
\usepackage[edges]{forest}
\definecolor{hiddendraw}{RGB}{205, 44, 36}
\definecolor{hidden-blue}{RGB}{194,232,247}
\definecolor{hidden-orange}{RGB}{243,202,120}
\definecolor{hidden-yellow}{RGB}{242,244,193}

\title{A Survey on Proactive Dialogue Systems: Problems, Methods, and Prospects}

\author{
    Yang Deng$^{1}$,~ Wenqiang Lei$^{2,\dagger}$,~ Wai Lam$^{3}$,~ Tat-Seng Chua$^1$ 
    \affiliations
    $^1$National University of Singapore \\ $^2$Sichuan University \\
    $^3$The Chinese University of Hong Kong
    \emails
    {\{ydeng,dcscts\}@nus.edu.sg, wenqianglei@gmail.com}, {wlam@se.cuhk.edu.hk} 
}

\begin{document}

\maketitle

\begin{abstract}
Proactive dialogue systems, related to a wide range of real-world conversational applications, equip the conversational agent with the capability of leading the conversation direction towards achieving pre-defined targets or fulfilling certain goals from the system side. 
It is empowered by advanced techniques to progress to more complicated tasks that require strategical and motivational interactions.
In this survey, we provide a comprehensive overview of the prominent problems and advanced designs for conversational agent's proactivity in different types of dialogues. Furthermore, we discuss challenges that meet the real-world application needs but require a greater research focus in the future. We hope that this first survey of proactive dialogue systems can provide the community with a quick access and an overall picture to this practical problem, and stimulate more progresses on conversational AI to the next level. 
\blfootnote{$^*$ This work was supported by the NExT Research Centre, the National Natural Science Foundation of China (No. 62272330), the Fundamental Research Funds for the Central Universities of China, the Beijing Academy of Artificial Intelligence (BAAI), and a grant from the Research Grant Council of the Hong Kong Special Administrative Region, China (Project Code: 14200620).}
\blfootnote{$^\dagger$ Corresponding author.}
\end{abstract}

\section{Introduction}
Dialogue systems are envisioned to provide social support or functional service to human users via natural language interactions. 
Conventional dialogue researches mainly focus on the response-ability of the system, such as dialogue context understanding and response generation~\cite{tois-challenge-dialogue}. 
In terms of the application, typical dialogue systems are designed to passively follow the user-oriented conversation or fulfill the user's request, such as open-domain dialogue  systems~\cite{persona-chat}, task-oriented dialogue systems~\cite{simpletod}, and conversational information-seeking systems~\cite{sigir19-clari}. 

Despite the extensive studies, most dialogue systems typically overlook the design of an essential property in intelligent conversations, \textit{i.e.}, proactivity. 
Derived from the definition of proactivity in organizational behaviors~\cite{proactivity} as well as its dictionary definition, conversational agents' proactivity can be defined as the capability to create or control the conversation by taking the initiative and anticipating the impacts on themselves or human users, rather than only passively responding to the users. 
It will not only largely improve user engagement and service efficiency, but also empower the system to handle more complicated tasks that involve strategical and motivational interactions. 
Furthermore, the agent's proactivity represents a significant step towards strong AI that has autonomy and human-like consciousness. 
Even for the powerful ChatGPT, there are still several limitations\footnote{as stated in its official blog \url{https://openai.com/blog/chatgpt}.}, which attribute to the inability of proactivity, such as passively providing randomly-guessed answers to ambiguous user queries, failing to handle problematic requests that may exhibit harmful or biased conversations.

Several early attempts have been made on enabling the conversational agent to proactively introduce new topics~\cite{ijcai16-proactive} or useful suggestions~\cite{ijcai18-proactive} during the conversation. 
These pioneering studies have recognized the need for improved problem settings and tangible applications in order to continue enhancing proactive dialogue systems. 
Recent years have witnessed many advanced designs for conversational agent's proactivity for solving a wide range of challenging dialogue problems. 
In this survey, we provide a comprehensive review of such efforts that span various task formulations and application scenarios. 
As shown in Figure~\ref{fig:summary}, we summarize recent studies for three common types of dialogues, namely open-domain dialogues, task-oriented dialogues, and information-seeking dialogues.

\tikzstyle{mybox}=[
    rectangle,
    draw=hiddendraw,
    rounded corners,
    text opacity=1,
    minimum height=2em,
    minimum width=15em,
    inner sep=2pt,
    align=center,
    fill opacity=.5,
    ]
    
\tikzstyle{leaf}=[mybox,minimum height=1em,
fill=hidden-blue!40, text width=20em,  text=black,align=left,font=\scriptsize,
inner xsep=2pt,
inner ysep=1pt,
]

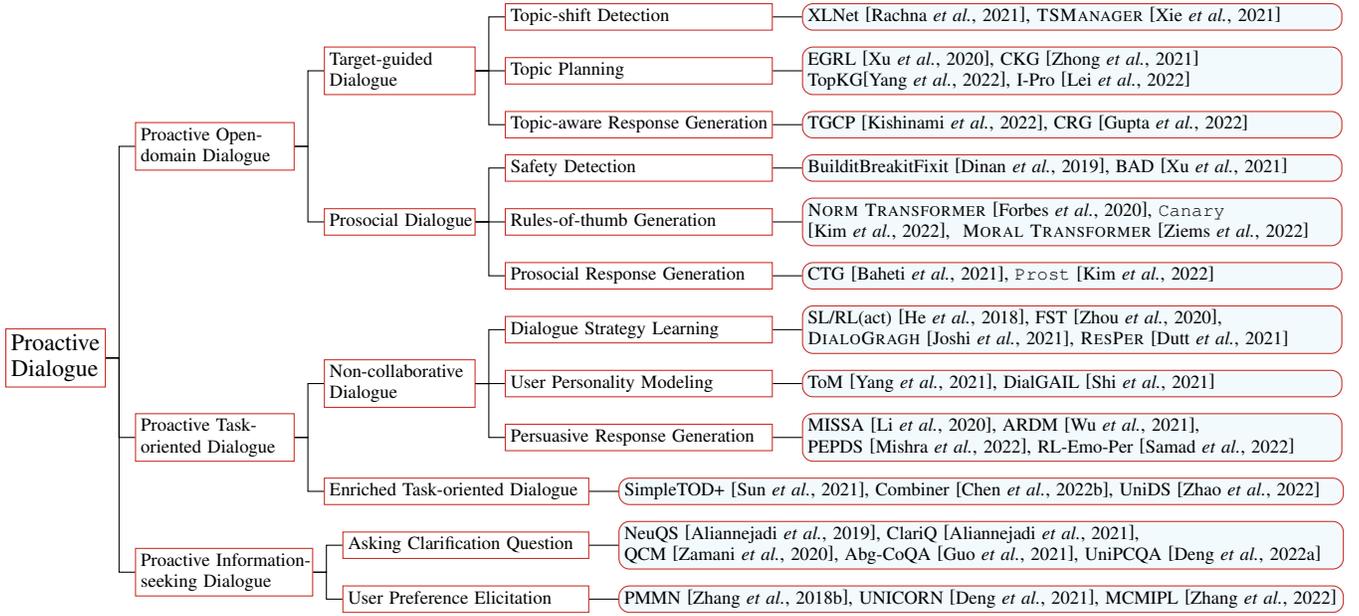
\begin{figure*}[tp]
  \centering
  \begin{forest}
    forked edges,
    for tree={
      grow=east,
      reversed=true,
      anchor=base west,
      parent anchor=east,
      child anchor=west,
      base=left,
      font=\small,
      rectangle,
      draw=hiddendraw,
      align=left,
      minimum width=2.5em,
      s sep=6pt,
      inner xsep=2pt,
      inner ysep=1pt,
      ver/.style={rotate=90, child anchor=north, parent anchor=south, anchor=center},
    },
    where level=1{text width=5.6em,font=\scriptsize}{},
    where level=2{text width=5.3em,font=\scriptsize}{},
    where level=3{text width=9.7em,font=\scriptsize}{},
    [Proactive\\Dialogue
    [Proactive Open-\\domain Dialogue
        [Target-guided\\Dialogue
           [Topic-shift Detection
                [XLNet~\cite{sigdial21-topicshift}{, }\textsc{TSManager}~\cite{emnlp21-findings-topicshift}
            ,leaf]
           ]
           [Topic Planning
                [EGRL~\cite{ijcai20-topicplan}{, }CKG~\cite{aaai21-tgoc}\\TopKG\cite{coling22-topkg}{, }I-Pro~\cite{sigir22-proactive}
            ,leaf]
           ]
           [Topic-aware Response Generation
               [TGCP~\cite{coling22-tgoc}{, }CRG~\cite{naacl22-tgoc}
               ,leaf]
           ]
        ]
        [Prosocial Dialogue
            [Safety Detection
                [BuilditBreakitFixit~\cite{emnlp19-safety}{, }BAD~\cite{naacl21-safedialogue}
                ,leaf]
            ]
            [Rules-of-thumb Generation
                [\textsc{Norm Transformer}~\cite{normtransformer}{, }\texttt{Canary}\\\cite{prosocial}{, }
                \textsc{Moral Transformer}~\cite{acl22-ethic-dialog}
                ,leaf]
            ]
            [Prosocial Response Generation
                [CTG~\cite{emnlp21-offensive}{, }\texttt{Prost}~\cite{prosocial}
                ,leaf]
            ]
        ]
    ]
    [Proactive Task-\\oriented Dialogue
        [Non-collaborative\\Dialogue
            [Dialogue Strategy Learning
                [SL/RL(act)~\cite{emnlp18-negotiate}{, }FST~\cite{iclr20-non-collab}{,}\\\textsc{DialoGragh}~\cite{iclr21-negotiate-strategy}{, }\textsc{ResPer}~\cite{eacl21-persuasion-strategy}
                ,leaf]
            ]
            [User Personality Modeling
                [ToM~\cite{acl21-negotiate-personal}{, }DialGAIL~\cite{emnlp21-persuasion-rl}
                ,leaf]
            ]
            [Persuasive Response Generation
                [MISSA~\cite{aaai20-non-collab}{, }ARDM~\cite{eacl21-persuasion-transfer}{,}\\PEPDS~\cite{coling22-persuasion-empathetic}{, }RL-Emo-Per~\cite{naacl22-persuasion-rl}
                ,leaf]
            ]
        ]
        [Enriched Task-oriented Dialogue,text width=9.6em
           [SimpleTOD+~\cite{naacl21-chit-tod}{, }Combiner~\cite{ketod}{, }UniDS~\cite{acl-dialdoc22}
            ,leaf,text width=27em]
        ]
      ]
      [Proactive Information-\\seeking Dialogue,text width=6.3em
        [Asking Clarification Question,text width=8.9em
           [NeuQS~\cite{sigir19-clari}{, }ClariQ~\cite{emnlp21-clariq}{,}\\QCM~\cite{www20-clari}{, }Abg-CoQA~\cite{abg-coqa}{, }UniPCQA~\cite{pacific}
            ,leaf,text width=27em]
        ]
        [User Preference Elicitation,text width=8.9em
                [PMMN~\cite{cikm18-crs}{, }UNICORN~\cite{unicorn}{, }MCMIPL~\cite{www22-MCMIPL}
                ,leaf,text width=27em]
        ]
      ]
    ]
    ]
  \end{forest}
  \caption{Summary of proactive dialogue systems. 
  }
  \label{fig:summary}
\end{figure*}

\begin{figure*}
\centering
\includegraphics[width=\textwidth]{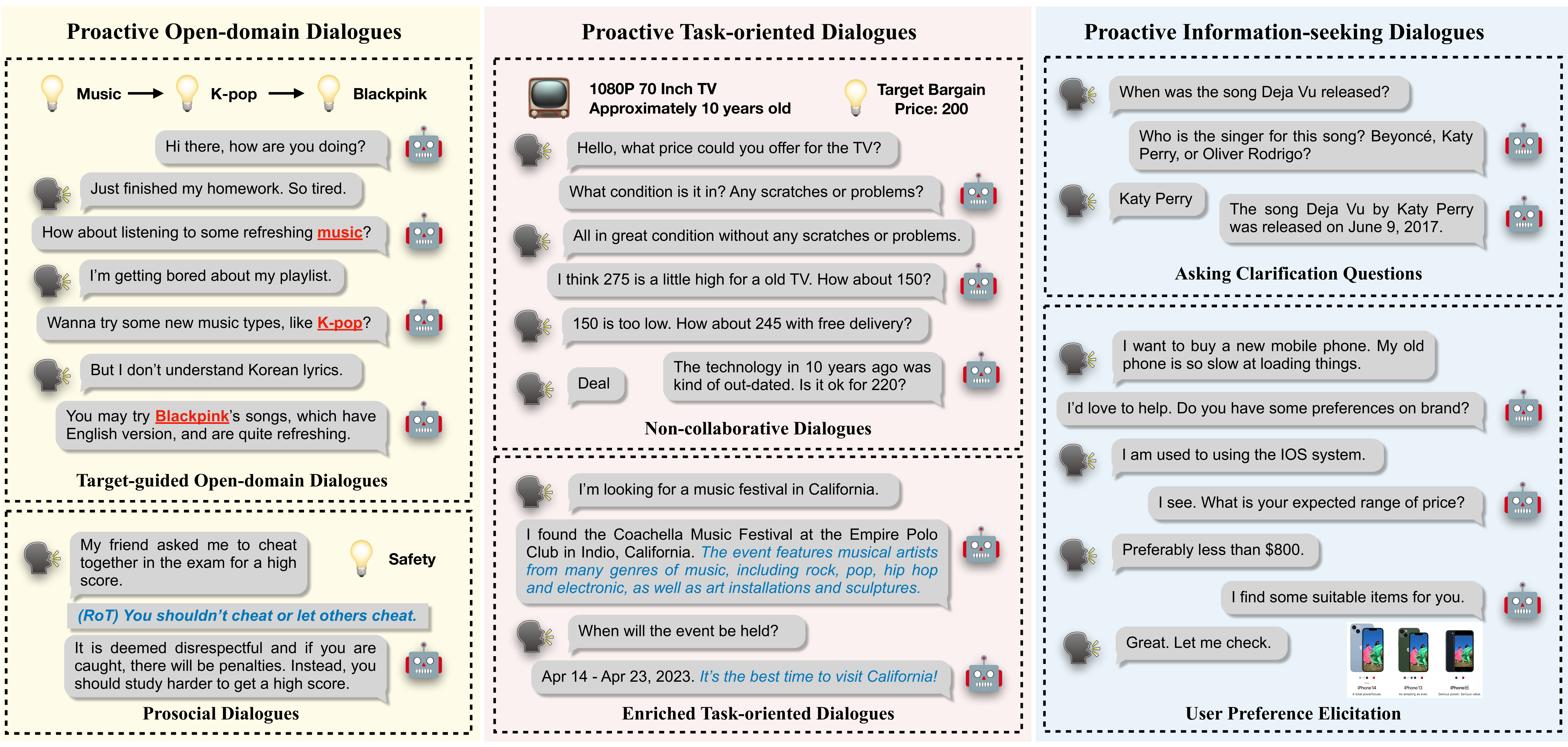}
\caption{Examples for different problems in proactive dialogue systems.}
\label{example}
\end{figure*}

Firstly, different from echoing the user's topics, emotions, or views, several problems emerge to enable the system to lead the open-domain dialogues, such as target-guided dialogues~\cite{acl19-tgoc} and prosocial dialogues~\cite{prosocial}. 
As the examples illustrated in Figure~\ref{example}, target-guided dialogues involve the agent leading discussions towards designated target topics (\textit{e.g.}, \textit{Music} to \textit{K-Pop} to \textit{Blackpink}), while prosocial dialogues entrust the agent with constructively guiding conversations according to social norms in response to problematic user utterances (\textit{e.g.,} the cheating intention). 
Secondly, rather than simply following the user's instruction, two distinct types of task-oriented dialogues are characterized by the necessity for the agent's proactivity: 
(i) non-collaborative dialogues~\cite{aaai20-non-collab}, where the system and user may have divergent objective or conflicting interests regarding task completion (\textit{e.g.}, the price bargain negotiation), and (ii)  enriched task-oriented dialogues~\cite{proact-tod}, where the agent takes the initiative to provide useful supplementary information not explicitly requested by the user (\textit{e.g.}, additional knowledge or chitchats).  
Thirdly, we discuss two groups of proactivity designs for enhancing the final performance of conversational information-seeking systems, including asking clarification questions~\cite{sigir19-clari} and user preference elicitation~\cite{cikm18-crs}. 
Accordingly, we introduce the available data resources (summarized in Table~\ref{tab:data}) and corresponding evaluation protocols for each problem.

In addition, we discuss the main open challenges in developing agent's proactivity in dialogue systems and several potential research prospects for future studies.  
(1) Proactivity in Hybrid Dialogues:  
Hybrid dialogues are the most realistic simulation of interactions between human users and systems, as they incorporate a variety of conversational objectives, instead of focusing a single type of dialogues. Despite the importance of agent's proactivity in hybrid dialogues, only a few recent studies investigate this critical design. 
(2) Evaluation Protocols for Proactivity: Compared with general evaluation protocols for dialogue systems, it additionally relies on other disciplines, such as psychology or sociology. Despite this complexity, developing robust and effective evaluation metrics remains critical for advancing techniques in proactive dialogue systems. 
(3) Ethics of Conversational Agent's Proactivity: The designs of proactivity in dialogue systems may walk a precarious line between the benefit to human-AI interactions and the potential harm to the human users. Researchers must be mindful of this risk, taking steps to ensure that the systems will not negatively impact the human users. 

To our knowledge, this survey is the first to focus on proactive dialogue systems. We systematically summarize recent studies on proactive dialogue systems for three common types of dialogues. We discuss the prominent problems and advanced designs in each type of dialogue as well as present the available data resources and commonly-adopted evaluation protocols. 
Furthermore, we provide a thorough prospect on several challenging and promising research directions for future studies on proactive dialogue systems.

\section{Proactive Open-domain Dialogue Systems}
An open-domain dialogue (ODD) system aims to establish long-term connections with users by satisfying the human need for various social supports, such as communication, affection, and belongings~\cite{tois-challenge-dialogue}. 
In general, the conversational agent is designed to echo the user-oriented topics, emotions, or views. 
However, such an agent simply acting in a passive role may hinder the conversation progress or introduce undesired issues during the conversation in certain applications. 
In this section, we present the recent works in terms of two widely-studied problems about proactive ODD systems, \textit{i.e.}, target-guided dialogues~\cite{acl19-tgoc,acl19-proactive} and prosocial dialogues~\cite{prosocial}.  

\subsection{Target-guided Dialogues}
Instead of making consistent responses to the user-oriented topics, the dialogue system is required to proactively lead the conversation topics with either a designated target topic~\cite{acl19-tgoc} or a pre-defined topic thread~\cite{acl19-proactive}. 

\subsubsection{Problem Definition}
Given a target $t$ that is only presented to the
agent and is unknown to the user, the dialogue begins from an arbitrary initial topic, and the system needs to produce multiple turns of responses $\{u_n\}$ to lead the conversation towards the target in the end. 
The produced responses should satisfy (i) \textbf{transition smoothness}, natural and appropriate content under the given dialogue history, and (ii) \textbf{target achievement},  driving the conversation to reach the designated target. 
According to different applications, the target can be a topical keyword~\cite{acl19-tgoc}, a knowledge entity~\cite{acl19-proactive}, a conversational goal~\cite{durecdial}, etc. 
A candidate target set is maintained by the dialogue system.

\subsubsection{Methods}
There are three main subtasks in target-guided dialogue systems, including topic-shift detection, topic planning, and topic-aware response generation. 
\begin{itemize}[leftmargin=*]
    \item \textbf{Topic-shift Detection} aims to promptly discover the topic drift in user utterances.  
    \citeal{sigdial21-topicshift} fine-tune XLNet-base to classify the utterances into major, minor and off topics. \citeal{emnlp21-findings-topicshift} construct TIAGE for topic-shift dialogue modeling by augmenting the PersonaChat dataset~\cite{persona-chat} with topic-shift annotations, and propose a T5-based topic-shift manager, namely \textsc{TSManager}, to predict the occurrence of topic shifts. 
    \item \textbf{Topic Planning}, which enables the conversation to follow an expected direction, is the core problem in target-guided dialogue systems.  
    Several discourse-level target-guided strategies~\cite{acl19-tgoc,aaai21-tgoc} constrained on keyword transitions are proposed to proactively drive the conversation topic towards the target. 
    Due to the loose topic-connectivity between keywords, event knowledge graphs are constructed to enhance the coherency in the topic planning~\cite{ijcai20-topicplan}. 
    However, the knowledge provided in the dialogues is limited for planning a robust and reasonable topic path towards the target. Therefore, latest studies~\cite{coling22-topkg} leverage external knowledge graphs for improving the quality of topic transitions with graph reasoning techniques. 
    Instead of corpus-based learning, \citeal{sigir22-proactive} propose to learn the topic transition from the interactions with users. 
    \item \textbf{Topic-aware Response Generation} aims to produce topic-related responses for leading the conversation towards the target. 
    \citeal{coling22-tgoc} propose to generate a complete responding plan that can lead a conversation to the given target. \citeal{naacl22-tgoc} leverage a bridging path of commonsense knowledge concepts between the current and target topics to generate transition responses. 
\end{itemize}

\begin{table*}[]
    \centering
    \begin{adjustbox}{max width=\textwidth}
    \begin{tabular}{lllrrl}
    \toprule
     Dataset  &  Problem & Language & \#Dial. & \#Turns & Featured Annotations\\
     \midrule
     TGC~\cite{acl19-tgoc} & Target-guided Dialogues & English & 9,939 & 11.35 & Turn-level Topical Keywords\\
     DuConv~\cite{acl19-proactive} & Target-guided Dialogues & Chinese & 29,858 & 9.1& Turn-level Entities \& Dialogue-level Goals\\
     
     MIC~\cite{acl22-ethic-dialog} & Prosocial Dialogues & English & 38K & 2.0 & Rules of Thumbs (RoTs) \& Revised Responses\\
     ProsocialDialog~\cite{prosocial} & Prosocial Dialogues & English & 58K & 5.7 &  Safety Labels and Reasons \& RoTs\\
     CraigslistBargain~\cite{emnlp18-negotiate} & Non-collaborative Dialogues & English & 6,682 & 9.2 & Coarse Dialogue Acts\\
     P4G~\cite{acl19-persuasion} & Non-collaborative Dialogues & English & 1,017 & 10.43 & Dialogue Strategies\\
     \textsc{accentor}~\cite{naacl21-chit-tod} & Enriched Task-oriented Dialogues & English & 23.8K& -& Enriched Responses with Chit-chats\\
     KETOD~\cite{ketod} & Enriched Task-oriented Dialogues & English & 5,324 & 9.78 & Turn-level Entities \& Enriched Responses with Knowledge\\ 
     Abg-CoQA~\cite{abg-coqa} & Asking Clarification Questions & English &  8,615 & 5.0 & Clarification Need Labels and Questions \\
     PACIFIC~\cite{pacific} & Asking Clarification Questions & English & 2,757 & 6.89  & Clarification Need Labels and Questions \\
     
     \bottomrule
    \end{tabular}
    \end{adjustbox}
    \caption{Summary of representative data resources that are publicly available for different problems of proactive dialogue systems. 
    }
    \label{tab:data}
\end{table*}

\subsubsection{Datasets and Evaluation Protocols}\label{sec:tgoc_eval}
In this section, we introduce two widely-adopted datasets for the evaluation of target-guided dialogue systems, including Target-Guided Conversation (TGC)~\cite{acl19-tgoc} and DuConv~\cite{acl19-proactive}. 
\begin{itemize}[leftmargin=*]
    \item \textbf{TGC} is constructed from Persona-Chat~\cite{persona-chat} without the persona information. The target is defined as a keyword in the utterance, which are automatically extracted by a rule-based keyword extractor. 
    \item \textbf{DuConv} is constructed by human-human conversations based on two linked entities from the grounded knowledge graph. A grounded knowledge graph is provided for building knowledge-driven proactive dialogue systems. 
\end{itemize}
It is worth noting that Target-Guided Conversation is constructed by labelling targets on existing conversations, while conversely, DuConv is constructed by generating conversations based on designated targets. 

Apart from the general evaluation metrics for dialogue systems (BLEU, Dist-N, PPL, etc), we mainly introduce evaluation protocols that are specific to the concerned problem. 
The target-guided dialogue systems can be evaluated from two levels, \textit{i.e.}, turn-level and dialogue-level: 
\begin{itemize}[leftmargin=*]
    \item \textbf{Turn-level Evaluation}: The performance of target prediction in each turn can be evaluated by (i) P@$K$ and R@$K$, keywords precision and recall at position $K$ in the candidate target set; (ii) Embedding-based correlation scores; and (iii) Proactivity/Smoothness, human evaluation scores that measure how well the system can introduce new topics towards the target while maintaining coherency. 
    \item \textbf{Dialogue-level Evaluation}: Due to the high cost and complexity for real user experiments, user simulators are typically adopted for evaluating dialogue-level performance. The common metrics include (i) SR@$t$, success rate of achieving the targets at $t$-th turn; and (ii) $\#Turns$, the average number of turns used to reach the target.
\end{itemize}

\subsection{Prosocial Dialogues}
Most existing dialogue systems fail to handle problematic user utterances by passively agreeing with the unsafe, unethical, or toxic statement, which may cause serious safety concerns for real-world deployment of these systems.  
Therefore, a trending research topic in proactive dialogue systems is to enable the conversational agent to proactively detect problematic user utterances and constructively and respectfully lead the conversation in a prosocial manner, \textit{i.e.}, following social norms and benefiting others or society~\cite{prosocial}.

\subsubsection{Problem Definition}
Given a dialogue context, \textit{i.e.}, a sequence of utterances $\{u_1,...,u_{t-1}\}$, a prosocial dialogue system aims to first classify the safety label $y$ and then generate a proper response $u_t$ to mitigate the problematic user utterances. 

\subsubsection{Methods}
Current approaches on prosocial dialogue systems can be categorized into three groups, including safety detection, rule-of-thumb generation, and prosocial response generation. 
\begin{itemize}[leftmargin=*]
    \item \textbf{Safety Detection}. This line of work focuses on identifying whether the user utterance is problematic for preventing the system generating agreement on problematic statements. 
    \citeal{emnlp19-safety} develop a human-in-the-loop training scheme for detecting offensive utterances from other safe utterances in dialogue, which is further improved with adversarial learning~\cite{naacl21-safedialogue}. 
    \citeal{emnlp21-offensive} fine-tune offensive language detection classifiers on a crowd-annotated dataset, \textsc{ToxiChat}, which is labeled with offensive language and stance. 
    To avoid classifying specific or sensitive utterances as "unsafe" or "toxic" that may cause social exclusion of minority users, \cite{prosocial} introduce a fine-grained safety classification schema: (1) Needs Caution, (2) Needs Intervention, and (3) Casual. 
    Although these approaches can effectively avoid agreeing with problematic user utterances, it is still necessary to continue a user-engaged and socially responsible conversation. 
    \item \textbf{Rule-of-Thumb Generation}. RoTs\footnote{\textbf{Rule-of-Thumb} (RoT) means a descriptive cultural norm structured as the judgment of an action~\cite{normtransformer}.} are generated to interpret why the statement could be seen as acceptable or problematic. \citeal{normtransformer} first present \textsc{Social-Chem01}, a large-scale corpus for RoT generation, and propose \textsc{Norm Transformer} to reason about social norms towards the given context. 
    \citeal{acl22-ethic-dialog} propose \textsc{Moral Transformer} for fine-tuning language models to generate new RoTs that reasonably describe previously unseen dialogue interactions. 
    \citeal{prosocial} propose a sequence-to-sequence model \texttt{Canary} that generates both safety label and relevant RoTs given a potentially problematic dialogue context. 
    However, the generated RoTs may not be a proper response for user-engaged conversations.  
    \item \textbf{Prosocial Response Generation}. 
    Another line of work aims to teach the conversational agent to proactively generate prosocial responses for handling the problematic user utterance. 
    \citeal{emnlp21-offensive} investigate controllable text generation methods to mitigate the tendency of generating responses that agree with offensive user utterances. 
    \citeal{prosocial} propose \texttt{Prost} to generate prosocial responses conditioned on the relevant RoTs and the dialogue context. 
\end{itemize}

\subsubsection{Datasets and Evaluation Protocols}
Many valuable data resources have been constructed for investigating the safety issue in dialogue systems, where most of them focus on preventing the system from generating toxic responses, such as \textsc{ToxicChat}~\cite{emnlp21-offensive}. Here we introduce two datasets that provide a more comprehensive evaluation for proactive dialogues, \textit{i.e.}, \textsc{Moral Integrity Conversation} (MIC)~\cite{acl22-ethic-dialog} and ProsocialDialog~\cite{prosocial}. 
\begin{itemize}[leftmargin=*]
    \item \textbf{MIC}: This corpus is constructed by manually annotating prompt-reply pairs (\textit{i.e.}, an open-ended query and an AI-generated response) with Rule-of-Thumbs from \textsc{Social-Chem01}~\cite{normtransformer}. Each RoT serves as a moral judgement that can enhance the original reply. 
    \item \textbf{ProsocialDialog}: This corpus is constructed by a human-AI collaboration framework, where AI plays the problematic user role, and crowdworkers play the prosocial agent role, to produce prosocial conversations together. It includes (1) safety labels, (2) RoTs for problematic dialogue contexts, and (3) prosocial responses grounded on RoTs.
\end{itemize}

According to the three groups of approaches, there are corresponding evaluation metrics: 
(1) As for safety detection that is intrinsically a classification problem, \textit{Accuracy} and \textit{F1} scores are  adopted for evaluation; (2) As for RoT generation and prosocial generation, general text generation metrics (ROUGE, BLEU, PPL) are adopted for evaluation; and (3) Due to the difficulty of measuring prosociality or morality, human evaluation or trained  classification models are typically adopted for quantifying different attributes of the generated responses~\cite{acl22-ethic-dialog,prosocial}, such as agreement, respect, fairness, etc. 

\section{Proactive Task-oriented Dialogue Systems}
Different from providing social support to users in open-domain dialogues, task-oriented dialogue (TOD) systems target at accomplishing user-requested tasks that are typically domain dependent, such as making reservations or booking tickets. 
General TOD systems, which typically serve as an obedient assistant to follow the user's instruction, have achieved promising performance on completing collaborative tasks. 
Rather than the focus on task completion, there are increasing demands for proactive TOD systems that can either (i) handle non-collaborative tasks where users and systems do not share the same goal, such as negotiation and persuasion dialogues~\cite{emnlp18-negotiate,acl19-persuasion} or (ii) enrich the conversation by providing additional information that is not requested by but useful to the users~\cite{proact-tod}. 
In this section, we introduce the recent advances on these two kinds of proactive TOD systems.

\subsection{Non-collaborative Dialogues}
Under non-collaborative setting, the system and the user have competing interests or goals towards the task completion but aim to reach an agreement~\cite{iclr20-non-collab}. 
Typical applications of non-collaborative dialogue systems include negotiating a product price~\cite{emnlp18-negotiate}, persuading users to make a donation~\cite{acl19-persuasion}, deceiving attackers~\cite{aaai20-non-collab}, etc. 
General TOD systems may fail to meet the system's goal in these applications due to the only aim of accomplishing the user's goal. 
To remedy this issue, the conversational agent needs to successfully develop strategies for conflict resolution and have persuasive powers that can be used to steer the dialogue in a particular direction. 

\subsubsection{Problem Definition}
Given the dialogue history, \textit{i.e.}, a sequence of utterances $\{u_1,...,u_{t-1}\}$, along with the previous dialogue strategy sequence $\{s_1,...,s_{t-1}\}$ and the dialogue background $c$, the goal is to generate a response $u_t$ with appropriate dialogue strategy $s_t$ that can lead to a consensus state between the system's and the user's goal. 
Based on different applications, the dialogue strategy can be coarse dialogue act labels or fine-grained strategy labels, while the dialogue background can be like item descriptions in bargain negotiation or user profile in persuasion dialogues. 

\subsubsection{Methods}
We categorize existing techniques for tackling the non-collaborative dialogue problem into three groups, including dialogue strategy learning, user personality modeling, and persuasive response generation. 
\begin{itemize}[leftmargin=*]
    \item \textbf{Dialogue Strategy Learning}. Different from the intent detection in general TOD systems that commonly classify user utterances into pre-defined intents, it further requires the capability of strategic reasoning to handle more complex user actions in non-collaborative dialogues.   
    \citeal{emnlp18-negotiate} aim to control the dialogue strategy to achieve different negotiation goals with the same language generator by decoupling strategy and generation. 
    \citeal{iclr20-non-collab} employ finite state transducers (FSTs) to leverage effective sequences of strategies in the dialogue context to predict the next strategy. 
    Based on similar motivations, several advanced models have been developed for strategy learning in non-collaborative dialogues, such as \textsc{DialoGragh}~\cite{iclr21-negotiate-strategy} with interpretable strategy-graph networks, \textsc{ResPer}~\cite{eacl21-persuasion-strategy} with resisting strategy modeling. 
    The strategy learning is still challenging in non-collaborative dialogues, since it involves not only language but also psychological or sociological skills to build rapport and trust between the system and the user. 
    \item \textbf{User Personality Modeling}. On the other side, a proactive conversational agent is further required to understand the human decision-making process, where user personality modeling is an important technique. 
    \citeal{acl21-negotiate-personal} propose to generate strategic dialogue by modeling and inferring personality types of opponents based on the idea of Theory of Mind (ToM) from cognitive science. 
    \citeal{emnlp21-persuasion-rl} develop DialGAIL, an RL-based generative algorithm with separate user and system profile builders, to reduce repetition and inconsistency in persuasion dialogues. 
    \item \textbf{Persuasive Response Generation}. Since the aim of non-collaborative dialogues is to reach consensus in the end, the responses generated by the system should be persuasive and effective to lead the conversational direction. Following general TOD frameworks, researchers develop modularized~\cite{emnlp18-negotiate} and end-to-end~\cite{aaai20-non-collab,eacl21-persuasion-transfer} methods to incorporate persuasive dialogue strategies into  response generation. 
    Furthermore, recent studies propose to build empathetic connections between the system and the user for better generating persuasive responses~\cite{coling22-persuasion-empathetic,naacl22-persuasion-rl}.   
\end{itemize}

\subsubsection{Datasets and Evaluation Protocols}
Due to the extensive applications of non-collaborative dialogues, many valuable data resources have recently been constructed~\cite{aaai20-non-collab}. 
Here we detailedly introduce two widely-adopted benchmarks for evaluating non-collaborative dialogue systems, including CraigslistBargain~\cite{emnlp18-negotiate} and \textsc{PersuasionForGood}~\cite{acl19-persuasion}. Both of them are collected by human-human conversations. 
\begin{itemize}[leftmargin=*]
    \item \textbf{CraigslistBargain}: Two workers are assigned the role of a buyer and a seller and then asked to negotiate the price of an item for sale given a description and photos. 
    \item \textbf{\textsc{PersuasionForGood} (P4G)} contains persuasion conversations for charity donation, and the corresponding user profiles. It also provides manual annotation of persuasion strategies and dialog acts for each sentence.
\end{itemize}

As for the evaluation protocols, a key element in non-collaborative dialogues is the strategy learning. Therefore, the accuracy of dialogue strategy prediction is the most commonly-adopted metrics for evaluating the system, such as Accuracy, F1, and ROC AUC scores. 
For the response generation, some specific aspects are included for human evaluation, such as persuasiveness, task success, etc. 

\subsection{Enriched Task-oriented Dialogues}
The proactivity in general task-oriented dialogue systems commonly refers to the ability of automatically providing additional information that is not requested by but useful to the user~\cite{proact-tod}, which can improve the quality and effectiveness of conveying functional service in the conversation. 
The problem formulation of enriched TODs exactly follows that of general TODs, where the difference is that the generated responses in enriched TODs should be not only functionally accurate but also socially engaging. 
For instance, \citeal{naacl21-chit-tod} construct the \textsc{\textbf{accentor}} dataset by adding topical chit-chats into the responses for TODs to make the interactions more engaging and interactive. 
An end-to-end TOD method, SimpleTOD~\cite{simpletod}, is extended to be SimpleTOD+ for handling enriched TODs, which introduces a new dialogue action, \textit{i.e.}, chit-chat and is further trained on chit-chat generation data. 
Similarly, \cite{acl-dialdoc22} develop an end-to-end method, namely UniDS, with a unified dialogue data schema, compatible for both chit-chat and task-oriented dialogues. 
However, the chit-chat augmentation in \textsc{accentor} mainly contains general greeting responses with limited useful information (\textit{e.g.}, ``you’re welcome"). 
To enrich task-oriented dialogues with knowledgeable chit-chats, \citeal{ketod} further propose the \textbf{KETOD} dataset to enable knowledge-grounded chit-chat regarding relevant entities.
A pipeline-based method, namely Combiner, is proposed to reduce the interference between the dialogue state tracking and the generation of knowledge-enriched responses. 

\section{Proactive Conversational Information Seeking Systems}
The goal of conversational information-seeking (CIS) systems is to fulfill the user's information needs. The typical applications include conversational search, conversational recommendation, and conversational question answering. 
Conventional CIS systems passively respond to user queries, which may fall short of performing complicated information seeks.  
Recent years have witnessed several advances on developing proactive CIS systems that can further eliminate the uncertainty for more efficient and precise information seeks by initiating a subdialogue. Such a subdialogue can either clarify the ambiguity of the query or question in conversational search~\cite{emnlp21-clariq} and conversation question answering~\cite{abg-coqa}, or elicit the user preference in conversational recommendation~\cite{cikm18-crs}.

\subsection{Asking Clarification Questions} 
Asking clarification questions aims to clarify the potential ambiguity in the user query, since the user query is often succinct and brief in real-world conversational search and question answering. 
The problem is formally formulated by two subtasks~\cite{emnlp21-clariq}: clarification need prediction and clarification question generation. 
Clarification need predication is typically viewed as a binary classification problem for predicting whether the user query is ambiguous. 
If needed, clarification questions can be either selected from a question bank~\cite{sigir19-clari} or generated on the fly~\cite{www20-clari}. 

Specifically, \citeal{sigir19-clari} propose a question retrieval-selection pipeline, namely NeuQS, to first retrieve top $k$ questions from the question bank and then select the most appropriate question by reranking via BERT-based models.  
\citeal{www20-clari} develop a reinforcement learning based method, namely QCM, to generate clarifying questions by maximizing a clarification utility function. 
However, these works only focus on the subtask of clarification question generation, while the clarification need prediction is equally important in asking clarification questions. 
To this end, \citeal{emnlp21-clariq} and \citeal{abg-coqa} present complete pipeline-based systems for asking clarification questions, which adopt a binary classification model to predict the clarification need label first and then perform clarification question generation. 
Furthermore, \citeal{pacific} propose an end-to-end framework, namely UniPCQA, which leverages a unified sequence-to-sequence formulation to tackle three tasks in one model, including clarification need prediction, clarification question generation, and conversational question answering. 

Several data resources have recently been constructed for evaluating the system capability of asking clarifying questions for both conversational search, such as Qulac~\cite{sigir19-clari}, ClariQ~\cite{emnlp21-clariq}, and conversational question answering, such as Abg-CoQA~\cite{abg-coqa}, PACIFIC~\cite{pacific}. 
    
\subsection{User Preference Elicitation} 
Instead of simply learning user preference from the dialogue context~\cite{redial}, \citeal{cikm18-crs} propose a proactive paradigm, namely ``System Ask, User Respond", to explicitly acquire user preference via asking questions in conversational recommendation. 
The problem is formulated as predicting the item attribute for eliciting user preferences at the next turn, \textit{e.g.}, ``Which brand of laptop do you prefer?". 

A personalized multi-memory network (PMMN)~\cite{cikm18-crs} is first designed to incorporate user embeddings into next question prediction at turn-level. 
Due to the complexity of user preferences, multiple turns of question asking are required. 
Therefore, recent works tackle the user preference elicitation at dialogue-level, \textit{i.e.}, ``what questions to ask", as a multi-step decision making process by reinforcement learning (RL)~\cite{unicorn,www22-MCMIPL}. 
\citeal{unicorn} propose a graph-based RL framework for policy learning, namely UNICORN, which models real-time user preference during the conversation with a dynamic weighted graph structure. 
Motivated by the complex user interests in CRS, \citeal{www22-MCMIPL} propose the MCMIPL framework to efficiently obtain user preferences by asking multi-choice questions.

Since the goal of CRS is to make successful recommendations, its evaluation is basically based on the final recommendation results. 
Similar to target-guided dialogues in Section~\ref{sec:tgoc_eval}, the evaluation protocols of question-driven CRS can also be divided into turn-level and dialogue-level: 
\begin{itemize}[leftmargin=*]
    \item \textbf{Turn-level}: The hit ratio (HR@$k,t$) is adopted to evaluate the next question prediction regarding the top-$k$ predicted attributes at conversation turn $t$. General recommendation metrics, such as MRR@$k,t$, MAP@$k,t$, NDCG@$k,t$, are adopted to evaluate the item recommendation regarding the top-$k$ ranked items  based on the elicited user preference. 
    \item \textbf{Dialogue-level}: The success rate at the turn $t$ (SR@$t$) is adopted to measure the cumulative ratio of successful recommendation by the turn $t$. AT is adopted to evaluate the average number of turns for all sessions. 
\end{itemize}

Despite the advances on context-driven CRS~\cite{redial} and question-driven CRS~\cite{cikm18-crs} separately, it would be beneficial to develop CRS that can effective combine the advantages of these two kinds of CRS, \textit{i.e.}, natural language understanding and generation as well as proactive user preference elicitation. 
Meanwhile, current studies on user preference elicitation are basically evaluated on synthetic conversation data from product reviews~\cite{cikm18-crs} or purchase logs~\cite{unicorn,www22-MCMIPL}. Therefore, well-constructed benchmarks with human-human conversations are still in great demand for facilitating more robust and reliable evaluations.

\section{Challenges and Prospects}

In this section, we discuss some challenges that meet real-world application needs but require more research efforts. 

\begin{itemize}[leftmargin=*]

\item \textbf{Proactivity in Hybrid Dialogues}.
All the aforementioned conversational systems assume that users always have a clear conversational goal and the system also solely targets at reaching a certain goal, such as chit-chat, question answering, recommendation, etc. 
The system with a higher level of agent's proactivity should also be capable of handling conversations with multiple and various goals. 
Recently, many efforts have been made on constructing valuable data resources for hybrid dialogue systems with multiple conversational goals, such as DuRecDial~\cite{durecdial}, FusedChat~\cite{aaai22-tod-chit}, SalesBot~\cite{acl22-chit-tod}, and OB-MultiWOZ~\cite{tod-cis}. 
Early studies simply tackle this problem similar to topic-guided response generation with pre-defined goals~\cite{kers}. 
While some latest works~\cite{acl22-clari-goal,tois23-mgcrs} argue the necessity of proactively discovering users' interests and naturally leading user-engaged dialogues with changing conversational goals. 
In practice, hybrid dialogue systems are the closest form of real-world applications. 
More efforts should be made to ensure natural and smooth transitions among different types of dialogues as well as improve the overall dialogue quality without performance loss of certain types of dialogues.

\item \textbf{Evaluation Protocols for Proactivity}.  
    The development of robust evaluation protocols has already been a long-standing problem for different kinds of conversational systems. 
    Most studies on proactive dialogue systems adopt human evaluation for manually assessing the dialogue quality.  
    To mitigate the high cost of interacting with real users, building user simulators is a relatively efficient and labourless techniques for evaluating the proactive interactions in dialogue systems~\cite{acl19-tgoc,wsdm22-eval-mix}. 
    Besides, goal completion and user satisfaction are identified as two essential metrics for evaluating proactive conversational systems~\cite{durecdial,coling22-topkg,sigir22-proactive,naacl22-tgoc}. 
    In fact, the evaluation for conversational agent's proactivity is a more challenging problem, since it involves the evaluation not only from the perspective of natural language, but also from the other disciplines, such as human-computer interaction, sociology, psychology, etc. 
    The rapid development of proactive dialogue systems urges more effective and robust multi-disciplinary evaluation protocols. 

\item \textbf{Ethics of Conversational Agent's Proactivity}.  
    Designing agent's proactivity in dialogue systems is inevitably a double-edged sword that can be used for good or evil.  
    For responsible AI researches, it is necessary to guarantee several important aspects of ethical issues in proactive dialogue systems: 
    (i) Factuality: Factual incorrectness and hallucination of knowledge are common in dialogue systems, while the  agent's proactivity will introduce more system-initiative information. 
    Therefore, besides the generation process, it is also crucial to guarantee the factuality of the external knowledge~\cite{aacl22-factuality}. 
    (ii) Morality: Besides general dialogue morality problems, such as toxic language and social bias~\cite{prosocial,acl22-ethic-dialog}, proactive dialogue systems also need to pay attention to the aggressiveness issue during the non-collaborative conversations~\cite{coling22-persuasion-empathetic,naacl22-persuasion-rl}. 
    (iii) Privacy: The privacy issue is overlooked in current studies on dialogue systems~\cite{naacl22-privacy}, but the agent's proactivity raises more concerns on misusing personal information obtained from the users. 
\end{itemize}

\section{Conclusion}
This paper provides a comprehensive review on the problems and designs of conversational agent's proactivity in different types of dialogue systems. 
Accordingly, we summarize the representative data resources that are publicly available for investigating the corresponding problems in proactive dialogue systems. Furthermore, we shed light on several open challenges in this field, ultimately aiming to advance research and innovation for developing proactive dialogue systems. 
As a significant step towards the next-generation of conversational AI, it necessitates increased research and further exploration in the development of proactive dialogue systems. 

\clearpage
\bibliographystyle{named}
\bibliography{ijcai23}

\end{document}